\documentclass{article}
\usepackage{times}
\usepackage{graphicx} 
\usepackage{subfigure} 
\usepackage{subfiles}
\graphicspath{{images}{../images/}}

\usepackage{natbib}
\usepackage{algorithm}
\usepackage{algorithmic}
\usepackage{amsmath}
\usepackage{amssymb} 
\usepackage{bm}
\usepackage{graphicx}
\usepackage{hyperref}
\usepackage{amsfonts}       
\usepackage{nicefrac} 

\newcommand{\vect}[1]{\boldsymbol{#1}}

 \DeclareMathOperator{\Tr}{Tr}
 
\usepackage[turnoff]{notes}

\usepackage[accepted]{icml2017}

\icmltitlerunning{Regularising Non-linear Models Using Feature Side-information}

\begin{document} 

\twocolumn[
\icmltitle{Regularising Non-linear Models Using Feature Side-information}

\icmlsetsymbol{equal}{*}
\begin{icmlauthorlist}
\icmlauthor{ Amina Mollaysa}{to}
\icmlauthor{Pablo Strasser}{goo}
\icmlauthor{Alexandros Kalousis}{ed}
\end{icmlauthorlist}

\icmlaffiliation{to}{HES-SO \& University of Geneva, Switzerland}
\icmlaffiliation{goo}{HES-SO \& University of Geneva, Switzerland}
\icmlaffiliation{ed}{HES-SO \& University of Geneva, Switzerland}

\icmlcorrespondingauthor{Amina Mollaysa}{maolaaisha.aminanmu@hesge.ch}
\icmlcorrespondingauthor{Pablo Strasser}{pablo.strasser@hesge.ch}
\icmlcorrespondingauthor{Alexandros Kalousis}{Alexandros.Kalousis@hesge.ch}
\icmlkeywords{feature side-information,  non-linear model}

\vskip 0.3in
]

\printAffiliationsAndNotice{}  

\begin{abstract} 
Very often features come with their own vectorial descriptions which provide detailed information about their properties. 
We refer to these vectorial descriptions as feature side-information. In the standard learning scenario, input is represented as a vector of features and the feature side-information is most often ignored 
or used only for feature selection prior to model fitting. We believe that feature side-information which carries information about features intrinsic property will help improve model prediction if used in a proper way during learning process. 
In this paper, we propose a framework that allows for the incorporation 
of the feature side-information during the learning of very general model families to improve the prediction performance. We control the structures of the learned 
models so that they reflect features' similarities as these are defined on the basis of the side-information. We perform 
experiments on a number of benchmark datasets which show significant predictive performance gains, over a number of baselines,
as a result of the exploitation of the side-information.
\end{abstract} 

\section{Introduction}
\label{Introduction}

Side-information in machine learning is a very general term used in very different learning scenarios with quite different
connotations. Nevertheless, generally it is understood as any type of information, other than the learning instances, which can be 
used to support the learning process; typically such information will live in a different space than the learning instances. Examples 
include learning with privileged information~\cite{JMLR:v16:vapnik15b} in which during training a teacher provides additional information 
for the learning instances; this information is not available in testing. In metric learning and clustering, it has been used to denote the
availability of additional similarity information on instances, i.e. pairs of similar and dissimilar instances,~\cite{DBLP:conf/nips/XingNJR02}.
In this paper we focus on side-information describing the features.
We will consider learning problems in which we have additional information describing the properties and/or the relations of the 
features. The features will have their own vectorial descriptions in some space in which we will describe their 
properties.

Real world problems with such properties are very common. For example in drug efficiency prediction problems, and more general 
in chemical formulae property prediction problems, drugs/formulae are collections of molecules. Each molecule comes with its own 
description, for example in terms of its physio-chemical properties, and/or its molecular structure. In language modeling, words are 
features and the words' semantic and syntactic properties are their side-information. In image recognition, pixels are features 
and their position is the side-information, and so on. Similar ideas also appear in tasks such as matrix completion, robust 
PCA and collaborative filtering \cite{rao2015collaborative, chiang2016robust, chiang2015matrix}. There one seeks low rank 
matrix decompositions in which the component matrices are constrained to follow relationships given by side-information 
matrices, typically matrices which contain user and item descriptors.

Despite the prevalence of such problems, there has been surprisingly limited work on learning with feature side-information.
\citeauthor{krupka2008learning}, \citeyear{krupka2008learning}, used the feature side-information to perform feature selection
as a preprocessing step prior to any modelling or learning. More interestingly  \cite{krupka2007incorporating} exploit the 
feature side-information directly within the learning process, by forcing features that have similar side-information to have
a similar weight within a SVM model. 
\note[Magda]{so doing the same as you?}
One can think of this as a model regularisation technique in which we force the model structure, i.e. the feature parameters, to 
reflect the feature manifold as this is given by the feature side-information. In the same work the authors also provide an ad-hoc 
way to apply the same idea for non-linear models, more precisely polynomials of low degree. 
\note[Magda]{so the above svm was only for linear features?}
However the solution that they propose 
requires an explicit construction of the different non-linear terms, as well as appropriate definitions of the feature side-information 
that is associated with them. These definitions are hand-crafted and depend on the specific application problem. 
Beyond this ad-hoc approach it is far from clear how one could regularise general non-linear models so that they follow the feature manifold.

\note[Alexandros]{In the related work we need to discuss more closely \cite{krupka2008learning} and how they do what they do.
Shouldn't we also discuss the distributed feature learning of word2vec? since, if words are 
features, then we learn the feature side-information which in some sense is even more powerful 
than what we do.}

In this paper we present a method for the exploitation of feature side-information in non-linear models. The main idea is that
the learned model will treat in a similar manner features that are similar. Intuitively, exchanging the values of two very similar features
should only have a marginal effect on the model output. This is straightforward for linear models since 
we have direct access to how the model treats the features, i.e. the feature weights. In such a case one can design regularisers 
as \citeauthor{krupka2007incorporating}, \citeyear{krupka2007incorporating}, did which force the feature weights to reflect the feature manifold. 
An obvious choice would be to apply a Laplacian regulariser to the linear model, where the Laplacian is based on the feature similarity.
Such regularisers have been previously used for parameter shrinkage but only in the setting of linear models where one has direct access 
to the model parameters~\cite{10.2307/23033591}. However, in general non-linear models we no longer have access 
to the feature weights; the model parameters are shared between the features and we cannot disentangle them. 

We present a regulariser which forces the  learned model to be invariant/symmetric to relative changes in the values of similar 
features. It directly reflects the intuition that small changes in the values of similar features should have a small effect on
the model output. The regulariser relies on a measure of the model output sensitivity to changes in all possible pairs of features. 
The model sensitivity measure quantifies the norm of the change of the model output under all 
possible relative changes of the values of two features. We compute this norm by integrating 
over the relative changes and the data distribution. Integrating over the relative changes is problematic 
we thus give two ways to approximate the sensitivity measure.  In the first approach we rely on a first order 
Taylor expansion of the learned model under
which the sensitivity measure boils down to the squared norm of the difference of the partial derivatives of 
the model with respect to the input features. Under this approach the regulariser finally boils down 
to the application of a Laplacian regulariser on the Jacobian of the model. In the second approach we rely on
sampling and data augmentation to generate instances with appropriate relative changes over different feature 
pairs.  We approximate the value of the regulariser only on the augmented data.

We implement the above ideas in the context of neural networks, nevertheless it is relatively straightforward to use 
them in other non-linear models such as SVMs and kernels. 

\note[Alexandros]{The above might be too long. It can be shortened.}
\note[Magda]{yep agree, perhaps say something simple here as
we use 2 approaches to .... One using 1st order Taylor to approximate, ... 2nd uisng data augmentation.

Just explain in a few words what they do and keep the details for later.

Even for the sensitivity measure, I would make it simple, drop the norm and integrals.

Something like:
The regulariser relies on a measure .... The measure essentially quantifies ... (no mention of norm or integral here) . Due to unknown probl distrib (or what really makes it intractable) it is intractable and therefore we propose two approxiamtion methods. First based on 1st order Teylor, 2nd on data augmentation.

That's it. The rest goes later.}

We experiment on a number of text classification datasets in which the side-information is the word2vec representation of the words.
We compare against a number of baselines and we show significant performance improvements.

\note[Alexandros]{By the way why do we expect the regulariser to be useful? will it help learning better with less data? have
better learning rates?}
\note[Magda]{IMPORTANT!  I think I had similar comment last time}

\note[Alexandros]{In fact the regulariser is some kind of squared $\ell_2$ norm, which measures the total size of the model 
difference under all possible relative changes. Seeing it like that aren't there more meaningfull ways to control the $\lambda$s?}

\note[Alexandros]{Do we really need the differentiability assumption? The original regulariser does not use the derivative.}

\section{Learning Symmetric Models with Respect to Feature Similarity}
We consider supervised learning settings in which, in addition to the classical data matrix $\mathbf X: n \times d$ 
containing $n$ instances and $d$ features, and the target matrix $\mathbf{Y}: n \times m$, we are also given a matrix 
$\mathbf Z: d \times c$, the $i$th row of which, denoted by $\mathbf z_i$, contains a description of the $i$th 
feature. We call $\mathbf{Z}$  the feature side-information matrix.  Note how the $\mathbf Z$ matrix is fixed and independent 
of the training instances. 
As in the standard supervised setting, instances, $\mathbf x_i \in \mathcal X \subseteq \mathbb R^{d}$, are drawn i.i.d 
from some non-observed probability distributions $P(\mathcal X)$ and targets, 
$\mathbf y_i \in \mathcal Y \subseteq R^{m}$,
are assigned according to some non-observed 
conditional distribution $P(\mathcal Y|\mathcal X), \mathcal Y \in \mathbb R^{m}$.
\note[Removed]{As in the standard supervised setting, instances are drawn i.i.d from some non-observed
density function $P(\mathbf x), \mathbf x \in \mathbb R^{d}$ and targets are assigned according to some non-observed conditional density 
$P(\mathbf y|\mathbf x), \mathbf y \in \mathbb R^{m}$. }
\note[Magda]{Probablity distribution. Careful, not all prob distribs even have a density fucntion. Also notation:
\begin{itemize}
	\item P(X) if you speak about the prob distrib as a function, 
	\item P(x)  which is the same as P(X=x) if you speak about its evalution (which in fact is rarely, I guess).
	\item Same as funcitons. P(X) is f; P(x) is f(x)
\end{itemize}}
In the standard setting we learn a mapping from the input 
to the output $\vect \phi :  \mathbf{x}\in \mathbb{R}^d\rightarrow \mathbf{y}\in \mathbb{R}^m$ 
using the $\mathbf{X}, \mathbf{Y}$ matrices by optimizing some loss function $L$. 
In this paper we learn the input-output mapping using in addition to the 
$\mathbf{X}, \mathbf{Y}$, matrices the feature side-information $\mathbf{Z}$ matrix. 

We bring the feature side-information in the learning process through the feature
similarity matrix $\mathbf S \in \mathbb{R}^{d \times d}$ which we construct from $\mathbf Z$ as follows. 
Given two features $i,j$, with $\mathbf z_i, \mathbf z_j$, side-information vectors 
the $S_{ij}$ element of $\mathbf S$ contains their similarity given by some similarity 
function. We will denote by $\mathbf L=\mathbf D - \mathbf S$ the Laplacian of the similarity matrix $\mathbf S$;
$\mathbf D$ is the diagonal degree matrix with $D_{ii}=\sum_j S_{ij}$.

We use the similarity and the Laplacian matrices to constraint the learned model 
to treat in a similar manner features that have similar side-information. This is relatively
straightforward with linear models such as $\mathbf W \mathbf X^\text{T}, \mathbf W \in \mathbb R^{m \times d}$,
and can be achieved through the introduction of the Laplacian regulariser 
$\Tr{(\mathbf W \mathbf L \mathbf W^\text{T})}=\sum_{ij} ||\mathbf W_{.i} - \mathbf W_{.j}||^2 S_{ij}$ 
in the objective function where $\mathbf W_{.i}$ is the $i$th column vector of $\mathbf W$ 
containing the model parameters associated with the $i$th feature,~\cite{10.2307/23033591}. 
The Laplacian regulariser forces the parameter vectors of the features to cluster according to the feature similarity.

\begin{figure}
  \includegraphics[scale=0.25]{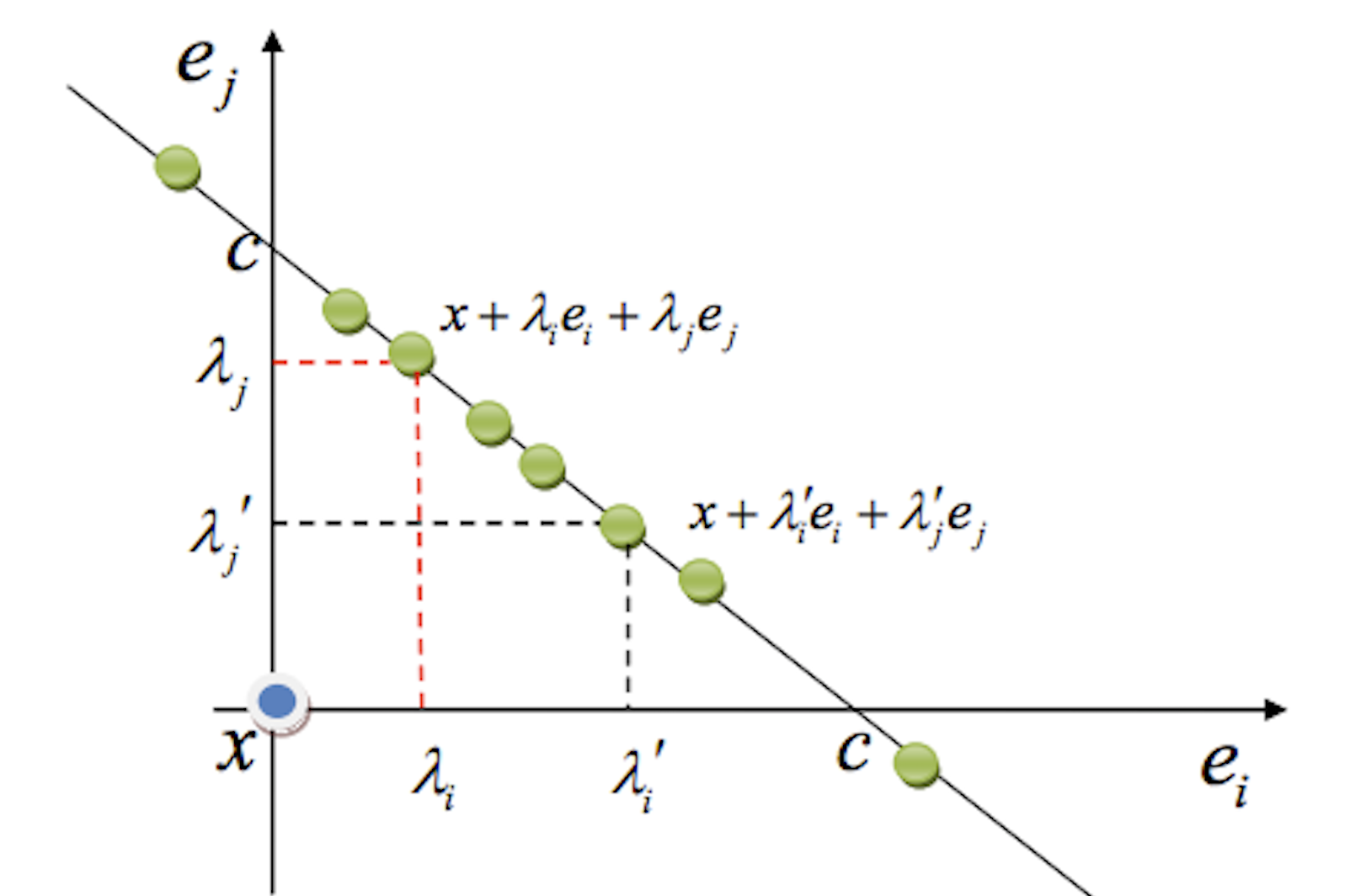}
  \caption{The blue dot is some given instance, $\mathbf x$. The two axes are the $i$th and $j$th features. 
	If the two features are on the limit identical then the model's output is constant along the line defined as: 
	$\mathbf x +  \lambda_i \mathbf e_i +  \lambda_j \mathbf e_j, \forall \lambda_i+\lambda_j=c$, where $c$ is some 
	constant.
	}
  \label{fig:mesh1}
\note[Magda]{
So initially the instance has zero of the two features? - the blue dot?
And you add to it both in the total amount of c but different proportions?
Or the instance already had a c in total and you're just changing the proportions?}
\end{figure}

However in non-linear models such neat separation of the model parameters is not possible since these are shared 
between the different input features. In order to achieve the same effect we will now operate directly on the model
output. We will do so by requiring that the change in the model's output is marginal if we change the relative 
proportion of two very similar features. Concretely, let $i$ and $j$ be such features, and $\mathbf e_i$, 
$\mathbf e_j$, be the $d$-dimensional unit vectors with the $i$th and $j$th dimensions respectively equal to one. We want that:
\begin{eqnarray}
\label{eq:nonLinearModelConstraint}
\vect \phi(\mathbf x + \lambda_i \mathbf e_i + \lambda_j \mathbf e_j) \approx 
\vect \phi(\mathbf x + \lambda_i' \mathbf e_i + \lambda_j' \mathbf e_j)
\end{eqnarray}
\begin{eqnarray*}
	\forall  \lambda_i,\lambda_j, \lambda_i',\lambda_j' \in \mathbb R 
	\text{ such that }\lambda_i + \lambda_j = \lambda_i' + \lambda_j'\nonumber \ 
\end{eqnarray*}
Equation~\eqref{eq:nonLinearModelConstraint} states that as long as the total contribution of the $i$, $j$, features is kept fixed, 
the model's output should be left almost unchanged. The exact equality will hold when the $i$, $j$, are on the limit identical, 
i.e. $S_{ij} \to \infty$.  More general the level of the  model's change should reflect the similarity of the $i,j,$ features, 
thus a more accurate reformulation of equation~\ref{eq:nonLinearModelConstraint} is:
\begin{eqnarray}
\label{eq:Constraint1}
||\vect \phi(\mathbf x+ \lambda_i \mathbf e_i + \lambda_j \mathbf e_j) - \vect \phi(\mathbf x + \lambda_i' \mathbf e_i + \lambda_j' \mathbf e_j)||^2
\propto \frac{1}{S_{ij} }
\end{eqnarray}
\begin{eqnarray*}
	\forall  \lambda_i,\lambda_j, \lambda_i',\lambda_j' \in \mathbb R 
	\text{ such that }\lambda_i + \lambda_j = \lambda_i' + \lambda_j'\nonumber \ 
\end{eqnarray*}
\note[Alexandros]{Do we have any issues by the fact that the maximum similarity is 1 and thus we fix the output difference to be 1 when the two features are identical?}
\note[Alexandros]{In addition I still have my doubts on whether the $\lambda$s should not be positive, and also the same goes for the feature values.}
\note[Amina]{feature values can be negative, lambdas can also be negative,  or we can translate the instance matrix by adding a positive big enough scalar and make all its value positive. }
Thus the norm of the change in the model output, that we get when we alter the relative proportion of two features $i$ and $j$, while 
keeping their total contribution fixed, should be inversely proportional to the features similarity, i.e. large similarity, small output change. 
The result is that the model is symmetric to similar features and its output does not depend on the individual contributions/values 
of two similar features but only on their total contribution. In figure \ref{fig:mesh1} we visualise the effect of the model 
constraint given in eq.~\ref{eq:Constraint1}. Given some instance $\mathbf x$ and two features $i$, $j$, that are on the limit 
identical the constraint forces the model output to be constant on the line defined by 
$\mathbf x + \lambda_i \mathbf e_i + \lambda_j \mathbf e_j$, $\forall \lambda_i+\lambda_j=c$, for some given $c \in \mathbb R$. 
We can think of the whole process as the model clustering together, to some latent factor, features that have very high 
similarity. The latent factor captures the original features total contribution leaving the model's output unaffected to relative changes 
in their values.

To unclutter notation we will define the vector $\vect \lambda = (\lambda_i,\lambda_j, \lambda_i',\lambda_j')$. We want 
the constraint of eq.~\ref{eq:Constraint1} to be valid over all instances drawn from $P(\mathbf x)$ as well as for all 
$\vect \lambda$ vectors that satisfy the equality constraint eq~\ref{eq:Constraint1}. A natural measure of the degree to which the constraint holds 
for the feature pair $i,j$ is given by:
\begin{eqnarray}
	\label{eq:constraintScore}
	R_{ij}(\vect \phi) & = &
	\int ||\vect \phi(\mathbf x + \lambda_i \mathbf e_i + \lambda_j \mathbf e_j) \\
        & & - \vect \phi(\mathbf x + \lambda_i'\mathbf e_i + \lambda_j'\mathbf e_j)||^2 \nonumber \\
	& & S_{ij} \mathbf I(\vect \lambda) P(\mathbf x) d\vect{\lambda}d\mathbf x \nonumber  
\end{eqnarray}
where  $\mathbf I(\vect \lambda) = 1 \text{ if }  \lambda_i+\lambda_j=\lambda_i'+\lambda_j', 
\text{ and \:$0$ \:otherwise}$. Since we want to define a regulariser that accounts for all feature pairs and their similarities we simply have:
\begin{eqnarray}
	\label{eq:Regularizer} R(\vect \phi) = \sum_{ij} R^d_{ij} (\vect \phi )
\end{eqnarray}
Calculating the regularizer is problematic
due to the presence of the $\mathbf I(\vect \lambda)$ function that selects the $\vect \lambda$ subspace over which the integration
is performed. In the next two sections we will give two ways to approximate it. The first one will 
be analytical relying on the first order Taylor expansion of $\vect \phi(\mathbf x)$ and its Jacobian. The second one stochastic,
essentially performing data augmentation and defining a regularisation term along the lines of eq.~\ref{eq:Constraint1}.

\note[Removed]{
\subsection{An analytical approximation}
We will use the first order Taylor expansion of $\vect \phi(\mathbf x)$ to simplify the 
squared term in eq.~\ref{eq:constraintScore} by removing the $\vect \lambda$ variable.
We approximate the value of  
$\vect \phi(\mathbf x + \lambda_i \mathbf e_i + \lambda_j \mathbf e_j) $ at $\mathbf {x}$ by:
\begin{eqnarray*}
\vect \phi(\mathbf x+ \lambda_i \mathbf e_i + \lambda_j \mathbf e_j) 
	&\approx & \phi(\mathbf x) +  \mathbf J(\mathbf x) \nonumber(\lambda_i \mathbf e_i + \lambda_j \mathbf e_j)
\end{eqnarray*}
$\mathbf J (\mathbf x) \in \mathbb R^{m \times d}$ is the Jacobian of $\vect \phi(\mathbf x)$ evaluated at $\mathbf x$.
Then using the Taylor expansion the squared term of eq~\ref{eq:constraintScore} becomes:
\begin{eqnarray}
	 \label{eq:jacobianApprox}
	 ||(\lambda_i - \lambda_i') \mathbf J (\mathbf x) \mathbf e_i -  (\lambda_j' - \lambda_j) \mathbf J (\mathbf x) \mathbf e_j||^2 
\end{eqnarray}
and since $\lambda_i + \lambda_j = \lambda_i' + \lambda_j'$ we have 
$(\lambda_i - \lambda_i')=(\lambda_j' - \lambda_j)$ and eq~\ref{eq:jacobianApprox} becomes:
\begin{eqnarray}
	\label{eq:jacobianApprox2}
||(\lambda_i - \lambda_i')(\mathbf J (\mathbf x)\mathbf e_i  - \mathbf J (\mathbf x) \mathbf e_j)||^2 = (\lambda_i - \lambda_i')^2||\nabla_i \vect \phi(\mathbf x) - \nabla_j \vect \phi(\mathbf x)||^2
\end{eqnarray}
where $\nabla_i \vect \phi(\mathbf x)$ is the $m$-dimensional partial derivative of $\vect \phi (\mathbf x)$ 
with respect to the $i$th input feature. Using eq~\ref{eq:jacobianApprox2} we can now approximate $R_{ij}$ as follows:
\begin{eqnarray*}
	\label{eq:constraintScoreApprox}
	R_{ij}(\vect \phi) & \approx &
\int (\lambda_i - \lambda_i')^2 d\vect{\lambda}	\int ||\nabla_i \vect \phi(\mathbf x) - \nabla_j \vect \phi(\mathbf x)||^2 S_{ij} P(\mathbf x) d\mathbf x \\
& \approx &\alpha \int ||\nabla_i \vect \phi(\mathbf x) - \nabla_j \vect \phi(\mathbf x)||^2 S_{ij} P(\mathbf x) d\mathbf x
\end{eqnarray*}
where $\alpha$ refers to a constant and can be merged with regularizer parameter later. $\alpha$
from which we get the following approximation of the $R(\vect \phi)$ regulariser:
\begin{eqnarray}
\label{eq:RegularizerApprox} 
	R(\vect \phi) & \approx &\sum_{ij} \int ||\nabla_i \vect \phi(\mathbf x) 
	                                                   - \nabla_j \vect \phi(\mathbf x)||^2 S_{ij} P(\mathbf x) d\mathbf x \nonumber \\
				& \approx & \int \sum_{ij} ||\nabla_i \vect \phi(\mathbf x) 
	                                                   - \nabla_j \vect \phi(\mathbf x)||^2 S_{ij} P(\mathbf x) d\mathbf x \nonumber \\
					& \approx & \int \Tr[\mathbf J(\mathbf x) \mathbf L \mathbf J^{\text{T}}(\mathbf x) P(\mathbf x) d\mathbf x
\end{eqnarray}
which is the local linear approximation of the original regulariser eq~\ref{eq:Regularizer}. 
Since we only have access to the training sample and not to $P(\mathbf x)$ we will get the sample estimate of 
eq.~\ref{eq:RegularizerApprox} given by
\begin{eqnarray}
	\label{eq:regulariserSampleEstimate}
	\hat{R}(\vect \phi) & = & \sum_{ij} \sum_k ||\nabla_i \vect \phi(\mathbf x_k) - \nabla_j \vect \phi(\mathbf x_k)||^2 S_{ij} \nonumber \\
				       & = & \sum_{k} \sum_{ij} ||\nabla_i \vect \phi(\mathbf x_k) - \nabla_j \vect \phi(\mathbf x_k)||^2 S_{ij} \nonumber \\
				       & = & \sum_{k}\Tr[\mathbf J(\mathbf x_k) \mathbf L \mathbf J^\text{T}(\mathbf x_k)]
\end{eqnarray}
So the sample based estimate of the regulariser is a sum of Laplacian regularisers applied on the Jacobian of each one of 
the training samples. It forces the partial derivatives of the model with respect to the input, or equivalently the model's 
sensitivity to the input features, to reflect the features similarity in the local neighborhood around each training point.
Or in other words it will constrain the learned model in a small neighborhood around each training point to have similar 
slope in the dimensions that are associated with similar features. 
Note that if $\vect \phi(\mathbf x)=\mathbf W \mathbf x$ then $\mathbf J(\mathbf x_k) = \mathbf W$ and 
$\Tr[\mathbf J(\mathbf x_k) \mathbf L \mathbf J^\text{T}(\mathbf x_k)]$ reduces to the standard $\Tr[\mathbf W 
\mathbf L \mathbf W^\text{T}]$ Laplacian regulariser on the columns of $\mathbf W$ associated with the input features.
Adding the sample based estimate of the regulariser to the loss function we get the final objective function which we 
minimize with $\vect \phi(\mathbf x)$ giving the following minimization problem under the analytical approximation:
\begin{equation}\label{Error function}
	\min_{\vect \phi} \sum_{k} L(\mathbf y_k,\mathbf \phi(\mathbf x_k))+\lambda \sum_{k}\Tr[\mathbf J(\mathbf x_k) \mathbf L \mathbf J^\text{T}(\mathbf x_k)]
\end{equation}
The approximation of the requlariser is only effective locally around each training point since it relies on first order
Taylor expansion. When the learned function is 
highly nonlinear, it can force model invariance only to small relative changes in the values of two similar features. However,
as the size of the relative changes increases and we move away from the local region the approximation is no longer effective. 
The regulariser will not be powerful enough to make the invariance hold away from the training points.  
If we want a less local approximation we can either use higher order Taylor approximation which is computationally prohibitive 
or rely on a more global approximation through data augmentation as we will see in the next section. Note also that the presence
of the Jacobian in the objective function means that if we optimise it using gradient descent we will need to compute second order
partial derivatives which come with an increasing computational cost. 
}
\note[Alexandros]{The above was removed because the transition from eq. \ref{eq:jacobianApprox} 
to eq. \ref{eq:jacobianApprox2} was problematic}

 \subsection{An analytical approximation}
We will use the first order Taylor expansion of $\vect \phi(\mathbf x)$ to simplify the 
squared term in eq.~\ref{eq:Constraint1} by removing the $\vect \lambda$ variable.
We will start by using the first order Taylor expansion to approximate the value of  
$\vect \phi(\mathbf x + \lambda_i \mathbf e_i + \lambda_j \mathbf e_j) $ at $\mathbf {x}$
\begin{eqnarray*}
\vect \phi(\mathbf x+ \lambda_i \mathbf e_i + \lambda_j \mathbf e_j) 
	&\approx & \phi(\mathbf x) +  \mathbf J(\mathbf x) \nonumber(\lambda_i \mathbf e_i + \lambda_j \mathbf e_j) \\
\end{eqnarray*}
$\mathbf J (\mathbf x) \in \mathbb R^{m \times d}$ is the Jacobian of $\vect \phi(\mathbf x)$ evaluated at $\mathbf x$.
Then plugging the Taylor expansion in eq~\ref{eq:Constraint1} we get:
\begin{eqnarray}
	 \label{eq:jacobianApprox}
	 ||(\lambda_i - \lambda_i') \mathbf J (\mathbf x) \mathbf e_i -  (\lambda_j' - \lambda_j) \mathbf J (\mathbf x) \mathbf e_j||^2 \propto \frac{1}{S_{ij}}
\end{eqnarray}
and since $\lambda_i + \lambda_j = \lambda_i' + \lambda_j'$ we have 
$(\lambda_i - \lambda_i')=(\lambda_j' - \lambda_j)$ and eq~\ref{eq:jacobianApprox} becomes:
\begin{eqnarray}
	\label{eq:jacobianApprox2}
||\mathbf J (\mathbf x)\mathbf e_i  - \mathbf J (\mathbf x) \mathbf e_j||^2 = ||\nabla_i \vect \phi(\mathbf x) - \nabla_j \vect \phi(\mathbf x)||^2 \propto \frac{1}{S_{ij}}
\end{eqnarray}
where $\nabla_i \vect \phi(\mathbf x)$ is the $m$-dimensional partial derivative of $\vect \phi (\mathbf x)$ 
with respect to the $i$th input feature. Using eq~\ref{eq:jacobianApprox2} we can approximate $R_{ij}$ as follows:
\begin{eqnarray*}
	\label{eq:constraintScoreApprox}
	R_{ij}(\vect \phi) & \approx &
	\int ||\nabla_i \vect \phi(\mathbf x) - \nabla_j \vect \phi(\mathbf x)||^2 S_{ij} P(\mathbf x) d\mathbf x 
\end{eqnarray*}
from which we get the following approximation of the $R(\vect \phi)$ regulariser:
\begin{eqnarray}
\label{eq:RegularizerApprox} 
	R(\vect \phi) & \approx &\sum_{ij} \int ||\nabla_i \vect \phi(\mathbf x) 
	                                                   - \nabla_j \vect \phi(\mathbf x)||^2 S_{ij} P(\mathbf x) d\mathbf x \nonumber \\
				& \approx & \int \sum_{ij} ||\nabla_i \vect \phi(\mathbf x) 
	                                                   - \nabla_j \vect \phi(\mathbf x)||^2 S_{ij} P(\mathbf x) d\mathbf x \nonumber \\
					& \approx & \int \Tr[\mathbf J(\mathbf x) \mathbf L \mathbf J^{\text{T}}(\mathbf x) P(\mathbf x) d\mathbf x
\end{eqnarray}
which is the local linear approximation of the original regulariser eq~\ref{eq:Regularizer} on the input instances.
Since we only have access to the training sample and not to $P(\mathbf x)$ we will get the sample estimate of 
eq.~\ref{eq:RegularizerApprox} given by
\begin{eqnarray}
	\label{eq:regulariserSampleEstimate}
	\hat{R}(\vect \phi) & = & \sum_{ij} \sum_k ||\nabla_i \vect \phi(\mathbf x_k) - \nabla_j \vect \phi(\mathbf x_k)||^2 S_{ij} \nonumber \\
				       & = & \sum_{k} \sum_{ij} ||\nabla_i \vect \phi(\mathbf x_k) - \nabla_j \vect \phi(\mathbf x_k)||^2 S_{ij} \nonumber \\
				       & = & \sum_{k}\Tr[\mathbf J(\mathbf x_k) \mathbf L \mathbf J^\text{T}(\mathbf x_k)]
\end{eqnarray}
So the sample based estimate of the regulariser is a sum of Laplacian regularisers applied on the Jacobian of each one of 
the training samples. It forces the partial derivatives of the model with respect to the input, or equivalently the model's 
sensitivity to the input features, to reflect the features similarity in the local neighborhood around each training point.
Or in other words it will constrain the learned model in a small neighborhood around each training point to have similar 
slop in the dimensions that are associated with similar features. 
Note that if $\vect \phi(\mathbf x)=\mathbf W \mathbf x$ then $\mathbf J(\mathbf x_k) = \mathbf W$ and 
$\Tr[\mathbf J(\mathbf x_k) \mathbf L \mathbf J^\text{T}(\mathbf x_k)]$ reduces to the standard $\Tr[\mathbf W 
\mathbf L \mathbf W^\text{T}]$ Laplacian regulariser on the columns of $\mathbf W$ associated with the input features.
Adding the sample based estimate of the regulariser to the loss function we get the final objective function which we 
minimize with $\vect \phi(\mathbf x)$ giving the following minimization problem under the analytical approximation:
\begin{equation}\label{Error function}
	\min_{\vect \phi} \sum_{k} L(\mathbf y_k,\mathbf \phi(\mathbf x_k))+\lambda \sum_{k}\Tr[\mathbf J(\mathbf x_k) \mathbf L \mathbf J^\text{T}(\mathbf x_k)]
\end{equation}
The approximation of the requlariser is only effective locally around each training point since it relies on first order
Taylor expansion. When the learned function is 
highly nonlinear, it can force model invariance only to small relative changes in the values of two similar features. However,
as the size of the relative changes increases and we move away from the local region the approximation is no longer effective. 
The regulariser will not be powerful enough to make the invariance hold away from the training points.  
If we want a less local approximation we can either use higher order Taylor approximation which is computationally prohibitive 
or rely on a more global approximation through data augmentation as we will see in the next section. Note also that the presence
of the Jacobian in the objective function means that if we optimise it using gradient descent we will need to compute second order
partial derivatives which come with an increasing computational cost.

\subsection{A stochastic approximation}
Instead of using the first order Taylor expansion to simplify the squared term required by the regulariser 
we can use sampling to approximate it. Concretely for a given
feature pair, $i,j$, and a given instance $\mathbf x$ we randomly sample $p$ quadruples $\lambda^{(l)}_i, \lambda^{(l)}_j, 
\lambda^{(l)'}_i, \lambda^{(l)'}_j \in \mathbb R$ such that $\lambda^{(l)}_i+\lambda^{(l)}_j= \lambda^{(l)'}_i+ \lambda^{(l)'}_j$, 
$l:=1 \dots p$, which we use to generate $p$ new instance pairs as follows:
\begin{equation*}
\mathbf x\rightarrow  \left\{
  \begin{array}{lr}
	  \mathbf x+ \lambda^{(l)}_i \mathbf e_i + \lambda^{(l)}_j \mathbf e_j\\
	  \mathbf x+ \lambda^{(l)'}_i \mathbf e_i + \lambda^{(l)'}_j \mathbf e_j 
  \end{array} \right.
  \end{equation*} 
We can now use the training sample and the sampling process to get an estimate of $R_{ij}(\vect \phi)$ by:
\begin{eqnarray*}
	& & \sum_k \sum_l  || ( \mathbf  \phi(\mathbf x_k + \lambda^{(l)}_i \mathbf e_i + \lambda^{(l)}_j \mathbf e_j) \\
	       & & -\mathbf \phi(\mathbf x_k + \lambda^{(l)'}_i \mathbf e_i + \lambda^{(l)'}_j \mathbf e_j)) ||^2S_{ij}
\end{eqnarray*}
and of the final regulariser $R(\vect \phi)$ by:
\begin{eqnarray}
	\label{eq:stochasticRegApprox}
	\tilde{R}(\vect \phi) & =&  \sum_{ij} \sum_k \sum_l || ( \mathbf  \phi(\mathbf x_k + \lambda^{(l)}_i \mathbf e_i + \lambda^{(l)}_j \mathbf e_j) \nonumber \\
	       & & -\mathbf \phi(\mathbf x_k + \lambda^{(l)'}_i \mathbf e_i + \lambda^{(l)'}_j \mathbf e_j)) ||^2S_{ij}
\end{eqnarray}
So the final optimization problem will now become:
\begin{eqnarray}\label{equation2}
\min_{\vect \phi}\sum_{k} L(\mathbf y_k,\mathbf \phi(\mathbf x_k))+ \lambda \tilde{R}(\vect \phi)
\end{eqnarray}
Note that the new instances appear only in the regulariser and not in the loss. The regulariser will penalise models which do not have 
the invariance property with respect to pairs of similar features. In practice when computing $\tilde{R}(\vect \phi)$
we do not want to go through all the pairs of features but only through the most similar. We do not want to spend 
sampling time on data augmentation for dissimilar pairs since for these there is no effective constraint on the 
values of the model's output. So we simplify the sum run only over the pairs of similar features.
One motivation for the stochastic approach was the fact that the analytical one relies in an approximation which is only effective
locally in the neighborhood of each learning instance. In the stochastic approach we have control on the size of the neighborhood
over which the constraint is enforced through the Euclidean norm of the change vector $(\lambda_i, \lambda_j)$; the larger its value the 
larger the neighborhood.
\note[Alexandros]{Now this is tricky, and this is why I think it is relevant whether $\lambda_i \geq 0$. If $\lambda_i=-\lambda_j$ 
that means that the size of the neighborhood is zero?}
\note[Magda]{agree with AK, tricky!
Also, if you pick lambdas and generate the perturb instances, how do you kow that 
these still live in the original X space and moreover how do you know that these are realistic instances under the P(X) ??}
The smaller the neighborhood the closer we are to the local behavior of 
the analytical approximation. We should note here that the sampling of stochastic approximation will naturally blend with the stochastic 
gradient optimization that we will use to optimize our objective functions.

\subsection{Optimization}

We learn $\vect{\phi}$ with a standard feed forward neural network with sigmoid activation functions 
applied on the hidden layers using stochastic gradient descent.

The objective function of the analytical approach contains the Jacobian of the model with respect to its input. Calculating the gradient 
over this results in the introduction of second order partial derivatives of the model with respect to the inputs and the model parameters.
\citeauthor{bishop1992exact}, \citeyear{bishop1992exact}, gave a backpropagation algorithm for the exact calculation of the Hessian of the 
loss of a multi-layer perceptron. We have adapted this algorithm so that we can compute the gradient of objective functions that contain
the Jacobian with respect to the input features, we give the complete gradient calculation procedure in the appendix.

\note[Amina]{The value of lambdas are not in R but at most in domain of 
features i , maybe it is better just $\Omega$.}
\note[Alexandros]{But the domain of the features is $\mathbb R$ so aren't we ok? See the definition we gave for $\mathbf x$
and the set to which it belongs.}

\note[Amina]{ The following can be removed, I was just trying to see if our iteration is enough to cover all similar pairs, 
which is not, and most of the case it even conver much earlie then the maximum iteration. } 

\note[Removed]{Ideally, if we want to cover all  $20\%$ most similar pairs, for a data set with feature dimention $d$, 
we need at least $\frac{{0.2d\times 
(0.2d-1)}}{2m}$ {\color{red} (this is not accurate in the case that same pairs might be repeated)} iteration, however, in most 
of the case, the objective function take less iteration than that to converge {\color{red} (and why this happen? because the 
rest of similarity is not useful??)}. }

We will give now the computational complexity of each the two methods. We will denote by $l$ the number of layers, 
$m$ the output dimension of the network, $h_k$ the number of hidden units of the $k$th layer and we will define 
$h_{\max}=\max\{h_k| k=1,\dots,l-1\}$.
The computational complexity of computing the gradient for a single instance 
of the objective function of the analytical approach is 
$\mathcal{O}( m\times h_1\times d^2)$ for networks with a single hidden layer and 
$\mathcal{O}(l\times m\times h_{\text{max}}^2\times d^2+l\times m\times h_{\text{max}}^3\times d)$ for networks 
with more than one hidden layers. 
\note[Magda]{I'm  no expert on algorithmic complexity but my impression was that the big O notation is used for asymptotic anlaysis.
That is my algo is O(n) as n goes to infinity. The other things are constant and therefore irrelevant.}
\note[Alexandros]{Magda is almost right. Two are the important things here $d$ and $n$; we have $d$ but I do not see $n$, I guess 
that that you are simply linear in $n$}
To reduce this computational complexity in our experiments we sparsify $\mathbf S$ 
by keeping only the entries correponding to top $20\%$ biggest elements and zero out the rest.  
\note[Alexandros]{This does not reduce the computational complexity, since $0.2*d$ is still on the order of $d$. 
To really reduce the computational complexity you should choose something that is independent of $d$ and small.}
\note[Magda]{how come that picking 20\% of S changed the d order in your complexities? What if you picked 50\%, 5\%, weird....}
The complexity now becomes $\mathcal{O}( m\times h_1\times d)$ for one layer networks and $\mathcal{O}(l\times m\times h_{\max}^2 
\times d+l\times m \times h_{\max}^3)$ for networks with more than one layer. The computational complexity of 
the stochastic approach is $\mathcal{O}( l \times h_{\max}^2 \times m\times p+l \times h_{\max} \times m \times p)$ 
while the computational complexity for standard feed forward network is  $\mathcal{O}( l \times h_{\max}^2 +l \times h_{\max} \times m)$.

\note[Alexandros]{This complexity $\mathcal{O}(l\times m\times h_{\max}^2 \times d+l\times m \times h_{\max}^3)$ means
that with more than one layers you probably need to significantly reduce your maximum number of hidden units.
Give also the computational complexity of the standrd NN for comparison reasons.}

\section{Related Work}

\note[Shrinked]
{\citeauthor{krupka2007incorporating}, \citeyear{krupka2007incorporating}, use feature side-information, they call it meta-features, 
within a linear SVM model. They force the SVM's feature weights to be similar for features that have similar side-information. 
They achieve that through the introduction of a Gaussian prior on the feature weight vector. The covariance matrix of the Gaussian
prior is given by the heat kernel over the features side-information, thus features with similar side-information will have a high 
covariance value. Under such a prior the SVM problem is equivalent to solving a standard SVM in a new space in which the feature weight 
correlations, given by the covariance of the prior, have been removed. This is also equivalent to solving a standard SVM problem where
instances are first transformed by the square root of the covariance matrix. The authors show how to apply the same idea to explicitly
computed polynomial functions, i.e. kernelisation. This requires also the explicit calculation of the meta-features for all monomials
of degree larger than one. This is done through a function of the meta-features of the features that participate in the monomial, which
which must be designed accounted for the requirements and particularities of the application domain. Obviously the approach does not 
scale with the number of features or with the degree of the polynomial and it is not applicable to nonlinear functions other than polynomials.
}

\citeauthor{krupka2007incorporating}, \citeyear{krupka2007incorporating}, use feature side-information, they call it meta-features, 
within a linear SVM model. They force the SVM's weights to be similar for features that have similar side-information. 
They achieve that through the introduction of a Gaussian prior on the feature weight vector. The covariance matrix of the Gaussian
is a function of the features similarity.
The authors show how to extend their approach from linear to polynomial models.
However, their approach requires explicit calculation of all the higher order 
terms limiting its applicability to low order polynomials.
Very similar in spirit is all the body of work on Laplacian regularisation for feature regularisation; 
\cite{krupka2007incorporating} contains an extensive review. Such regularisers constrain the feature
weights to reflect relations that are given by the Laplacian. The Laplacian matrix is constructed 
from available domain knowledge, what here we call feature side information. However, it can
also be constructed from the data; for example as a function of the feature correlation matrix.

\note[Shrinked]{
Our regulariser relies on a measure of the model's sensitivity to relative changes in the values of similar features. One way we 
quantify this sensitivity is through the model's Jacobian with respect to the input features. Regularisers on the Jacobian 
are used when one wants to stabilise/robustify models to noisy inputs. Relevant work includes
 contractive auto encoders, \cite{Rifai2011-ay},  manifold tangent classifiers, \cite{Rifai2011-hu}, and \cite{Zhai2015-gv}.
Such regularisers are typically matrix norms, e.g. Frobenius, of the Jacobian at the input instances. Their effect is to 
force the model to be relatively 
constant in small neighbors around the input instances, thus introducing invariance for small input variations. 
\citeauthor{Rifai2011-ay}, \citeyear{Rifai2011-ay}, distinguish between analytical and stochastic methods to model 
stabilisation, where the former rely on analytical approaches such as the Jacobian regularisation, and the latter 
on explicit input corruption and data augmentation. Denoising autoencoders, \cite{Vincent2010} follow the stochastic paradigm and 
require that small random variations in the inputs have only a limited effect on the model output. 
Optimizing the Jacobian in networks with more than one layer is cumbersome, thus very often the stochastic 
approach is prefered over the analytic e.g.~\cite{Zhai2015-gv}. Regularisers on higher order derivatives, Hessian,
are also used, \cite{DBLP:conf/pkdd/RifaiMVMBDG11}, in such cases the stochastic approach is the only choice due to 
the prohibitive cost of optimizing the Hessian term.
\citeauthor{DBLP:conf/cvpr/ZhengSLG16}, \citeyear{DBLP:conf/cvpr/ZhengSLG16},  also used Gaussian perturbations 
to stabilise the network's output with respect to variations in the input, 
essentially augmenting the training data. The authors note that {\em unlike} typical data augmentation approaches they do not evaluate 
the augmented instances with the loss function but only with the stability regulariser because they have observed that the opposite lead 
to underfitting.
}
\note[Alexandros]{Need to discuss the data augmentation for stability to noise in to more detail.
The manifold tangent classifiers, \cite{Rifai2011-hu}, uses the Jacobian of the last layer but it seems to be much
more elaborate than the other methods. Definetely to look at more closely.

Also you should take a close look on how these people use their regularisers, even though not of the same type as ours, 
do they combine them with other regularisers?
}

The Taylor expansion we use in the analytical approximation of the regulariser results in the use of the Jacobian 
of the model. Regularisers that use the Jacobian have previously been successfully used to control the stability/robustness 
of models to noisy inputs.  Relevant work includes contractive auto encoders, \cite{Rifai2011-ay},  
and \cite{Zhai2015-gv}. \cite{Rifai2011-ay} use the Frobenius of the Jacobian at the input instances 
to force the model to be relatively constant in small neighbors around the input instances. Such a 
regulariser introduces invariance to small input variations. In a different setting \citeauthor{Rosasco2013-lw}, \citeyear{Rosasco2013-lw},
used the Jacobian to learn sparse non-linear models in the context of kernels.

Optimizing the Jacobian in networks with more than one layer is cumbersome, thus very often the stochastic 
approach is preferred over the analytic e.g.~\cite{Zhai2015-gv}.  Denoising autoencoders, \cite{Vincent2010} 
follow the stochastic paradigm and require that small random variations in the inputs have only a limited 
effect on the model output. \citeauthor{DBLP:conf/cvpr/ZhengSLG16}, \citeyear{DBLP:conf/cvpr/ZhengSLG16},  
used Gaussian perturbations to stabilise the network's output with respect to variations in the input, 
essentially augmenting the training data. Regularisers on higher order derivatives, Hessian,
are also used, \cite{DBLP:conf/pkdd/RifaiMVMBDG11}, in such cases the stochastic approach is the only choice due to 
the prohibitive cost of optimizing the Hessian term.

\note[Shrinked]{
Input noise is one type of instance structure to which we know that the learned models should be 
invariant. Very often and depending on the type of application we have additional prior knowledge on the 
instance structures to which the models should be invariant. In such cases a typical approach is to generate 
training instances that exhibit such structures and let the model learn the invariances. For example for images 
such corruption structures include translations, rotations, scalings,etc, \cite{Simard1991, Decoste2002}. 
}
\note[Alexandros]{I think we are missing a lot here, i.e. on what kind of invariances people have sought 
to model through data augmentation or analytically. See for example the references in \cite{Vincent2010} section 4.2,
we should find more recent stuff in which we generate instances with specific structure.

Independently of that I also find very interesting the discussion in the same paper on the relation 
of Tikhonov regularisation to training with noise and how does/does-not that transfer to non-linear 
models, see also  ~\cite{Bishop:1995:TNE:211171.211185}.
}

Data augmentation is a well-established approach for learning models with built-in invariance to 
noise and/or robustness to data perturbations. In addition it is also used when we have additional 
prior knowledge on the instance structures to which the models should be invariant. In imaging problems
such structures include translations, rotations, scalings,etc, \cite{Simard1991, Decoste2002}.

\note[Shrinked]{
The Jacobian of the learned model with respect to the input features provides a handle to the model 
that allow us to apply regularisers that induce different effects and not only local invariance 
around the training instances. \citeauthor{Rosasco2013-lw}, \citeyear{Rosasco2013-lw}, use the 
Jacobian to perform feature selection for non-linear models, kernels in particular. They rely on the
simple observation that the partial derivative of the model with respect to a given feature captures 
the importance of that feature for the model. They use the $\ell_2$ 
norm of the derivative of the given feature computed over all instances to measure the importance of 
that feature and add this term as a regulariser in their objective function. The resulting regulariser
acts as a group lasso and forces many of the input features to have a zero derivative making thus the 
function flat along these dimensions and essentially performing feature selection. 
}

\note[Alexandros]{\cite{Zhai2015-gv} These guys they have a rather interesting comparison and discussion to other approaches. 
These include feature noising of~\cite{Bishop:1995:TNE:211171.211185} which adds noise to the features 
according to some distribution, essentially augmenting the instances. The augmented instances are evaluated 
in the loss function. In their discussion \cite{Zhai2015-gv} show that adding such noise (at least when the
noise $N(0,\sigma^2I)$ introduces a regulariser on the Hessian of the model at the inputs. This has as an
effect to control the curvature of the learned function but {\em does not} control the flatness of the learned 
model around the training instances. In the same discussion they also note that one variant of Dropout, 
~\cite{DBLP:journals/jmlr/SrivastavaHKSS14}, in which mask out noise is only applied to the input layer 
can be analysed in a similar manner and point readers to \cite{DBLP:conf/nips/WagerWL13} for further discussion.
For what is worth they also have an interesting discussion on the relation of their work to that of adversarial 
training~\cite{DBLP:journals/corr/GoodfellowSS14}. They note that NNs can missclassify examples which are 
generated by adding {\em certain} small perturbations on the learning examples, i.e. the adversarial examples, 
\cite{DBLP:journals/corr/SzegedyZSBEGF13}, to which adversarial learning has been proposed as a solution. 
Stabilising the network, whether analytically or stochastically, in the neighborhood of the training instances 
can also alleviate the effect of adversarial instances.}

\note[Removed]{
Work of \cite{rao2015collaborative} used 
user and item properties in the matrix completion task, similar to that, \cite{chiang2016robust} has used feature side-information in 
Robust PCA with the assumption that the underlying matrix is defined by an implicit bilinear function of  the features side-information.

In the work of \cite{bian2014knowledge}, syntactic and semantic knowledge of the word, which we can consider as feature side-information, 
has been used as additional input to a deep network to learn a more effective word embedding. 

Note that the regularizer we impose aims to lead the model to have
the same response on a pair of directions which is defined by the 
feature side-information in a specific region. The regularizer 
forces the function to respect the data structure. 

In the first approach, a laplacian regularuizer is applied on the columns 
of the jacobian of the learned function to let the learned function to have 
symmetric response to the similar features. 
}

\note[Removed]{Our regulariser enforces prior knowledge about feature relations derived from the feature side-information, constraining
the model to treat similar features in a similar manner, or in other words making it symmetric to similar features. Due 
to its intractability we approximate it using two different approaches, an analytical one and a stochastic one. In the 
analytical approach we apply a Laplacian regulariser on the Jacobian; we get the Laplacian from the feature similarities given by the feature side-information. This approach essentially
does Laplacian-based feature shrinkage for non-linear models.  In the stochastic approach we rely on instance augmentation, we generate 
instances according to the feature similarity relations, together with a term that constraints the model output sensitivity to reflect 
the feature similarity relations. 
}

The works that are closer to our work are \cite{krupka2007incorporating} as well as the 
works that use Laplacian based regularisers for model regularisation, e.g. ~\cite{10.2307/23033591}. However to the best of our knowledge all 
previous work was strictly limited to linear models.  We are the first ones who show how such regularisers and constraints can be applied 
to general classes of non-linear models.

\section{Experiments}
We will experiment and evaluate our regularisers in two settings, a synthetic and a real world one. We will compare
the analytical and the stochastic regulariser, which we will denote by AN and ST respectively, against popular regularisers 
used with neural networks, namely $\ell_2$ and Dropout \cite{srivastava2014dropout}, over different network 
architectures.  In the real world datasets we also give the results of the Word Mover's Distance, 
WMD, \cite{kusner2015word} which makes direct use of the side-information to compute document distances. 
Obviously our regularisers and WMD have an advantage 
over $\ell_2$ and dropout since it exploit side-information which $\ell_2$ and dropout do not.

We trained both the analytical and the stochastic models, as well as all baselines against which we compare,
using Adam \cite{kingma2014adam}. We used $\alpha=0.001,\beta_1=0.9, \beta_2=0.999$ for one hidden layer networks, 
and  $\alpha=0.001$ for the networks with more hidden layers. We initialize all networks parameters using 
\cite{glorot2010understanding}.
Due to the large computational complexity of the analytical approach we set the mini-batch size $m$ to five. 
For the stochastic model, as well as for all the baseline models, we set the mini-batch size to 20. 
For the analytical model we set the maximum number of iterations to 5000. 
For the stochastic model we set the maximum number of iterations to 10000 for the one layer networks and 
to 20000 for networks with more layers. We used early stopping where we keep 20\% of the training data as
the validation set. Every five parameter updates we compute the validation error. Training terminates either 
if we reach the maximum iteration number or the validation error keeps increasing more than ten times in a row.

In the stochastic approach we do the sampling for the generation of the instance pairs within 
the stochastic gradient descent process.  Concretely for each instance $\mathbf{x}$ in a mini 
batch we randomly chose a feature pair $i,j$, from the set of similar feature pairs. We sample 
a quadruple $\{\lambda_i, \lambda_j, \lambda_i',\lambda_j'\}$ from $\mathbb R
$ respecting the constraint: $\lambda_i+\lambda_j= \lambda_i'+\lambda_j'$ from which we generate the respective instance pairs. 
We repeat the process $p$ times each time sampling a new feature pair $i,j$, and a new quadruple.
We fix the set of similar feature pairs to be the top $20\%$ of most similar feature pairs. Thus
within each mini-batch of size $m$ we generate $m \times p$ instance pairs and we accumulate the 
norm of the respective model output differences in the objective. In the experiments we set $p=5$

\subsection{Artificial datasets}
We design a simple data generation process in order to test the performance of our regularisers when the
data generation mechanism is compatible with the assumptions of our models. We randomly 
generate an instance matrix $\mathbf{X}\in \mathbb{R}^{n\times d}$ by uniformly sampling instances from $\mathbb R^d$. 
We create $d/2$ feature clusters as follows. To each one of these clusters we initially assign one of the original 
input features without replacement. We assign randomly and uniformly the remaining $d/2$ features 
to the clusters. We use the feature clusters to define a latent space where every feature 
cluster gives rise to a latent feature. The value of each latent feature is the sum of the 
values of the features that belong to its cluster. On the latent space
representation of the training data we apply a linear transformation
that projects the latent space to a new space with lower dimensionality $q$. On this lower dimensionality space we apply an element-wise
sigmoid and the final class assignment is given by the index of the maximum sigmoid value. 
The similarity $S_{ij}$ of the $i,j$, features of the original space is 1 if they 
ended up in the same cluster and 0 otherwise. 
We set $d=3000$, $n=5000$, $q=5$. We will call this dataset A1.
The generating
procedure gave a very sparse $\mathbf S$ matrix with only $0.04\%$ of its entries 
being non-zero. Each feature had an average of 1.3 similar features.
We used 4000 instances for training and the rest for testing. During training 
20\% of the instances are used for the validation set. We measure performance 
with the classification error, i.e. percentage of wrong predictions.
We train all algorithms on the original input space.  For all regularisers we used 
a single layer with 100 hidden units. 
We tune the hyperparameters based on the performance on the validation set.  
We select the $\lambda$ hyperparameters of of AN, ST, and  $\ell_2$ from  $\{10^{k}|k=-3,\dots,3\}$;
we select the $\lambda$ of dropout from  $[0.1,0.2,0.3,0.4,0.5]$. We set the $c$ in the augmentation 
process, that controls the size of the neigborhood within which the output constraints should hold, 
to one.

\note[Alexandros]{Hhmmm this goes on the opposite direction of the discussion of how the sparsity
affects the performance. We have an extremely sparse matrix S which should mean that the regulariser should
have no effect. But still brings a great improvement. 

One issue with this generation procedure is that the algorithm is given perfect knowledge of the clustering information.
If we had worked from the side information to generate meaningful clusters and then use the side information to learn and
not directly the cluster information that would have been a more fair comparison. Which makes me think that it is this 
perfect cluster information that might explain the very good performance on the artificial data. Definitely we need some
more experiments here.
}

\begin{figure}
\begin{center}
\includegraphics[scale=0.25]{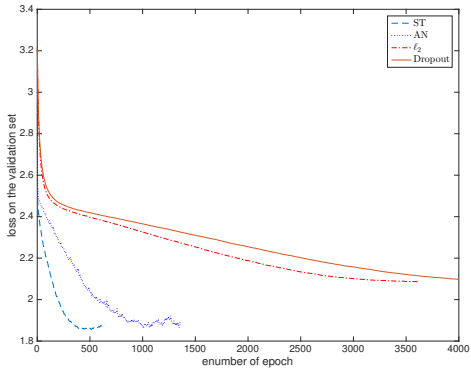}
\caption{\footnotesize{Learning curves for the different regularisers}}
\label{learning_curve}
\end{center}
\end{figure}

Both the analytical and the stochastic regulariser bring performance improvements of roughly 10\% 
when compared to the $\ell_2$ regulariser and to Dropout, results in table~\ref{artificial2}. In 
figure~\ref{learning_curve} we plot the learning curves, i.e. error on the validation set for each epoch, of the four regularisers. 
We can see that both the analytical and the stochastic regulariser converge much faster and to 
significantly lower error values than either $\ell_2$ or dropout.

\note[Alexandros]{In the figure change the labels of the lines to: AN, ST, $\ell_2$, 
Dropout. Use thicker lines and use also different line styles so that one can differentiate 
them even without colors. Remove the title on the top of the figure.}

\note[Alexandros]{I think we need additional and more elaborate artificial experiments. Two main directions.
Use more, and more complex, non-linearities
Use more complex ways to generate the similarities and the groupings. So far you only have pairs, I would like 
also to see experiments in which you say something like: I will now generate a feature cluster with $k$ features
where $k$ is chosen randomly. You proceed until all features have been selected and then you do the same thing
as you have doen here. Also would be interesting the see whether you can generated smoother similarities, not 
just 0/1.
}

The regularizer we propose constrains the model structure by forcing the model to reflect
the feature similarity structures as these are given in the similarity matrix. Thus we expect 
the structure of the similarity matrix to have an impact on the performance of the regulariser. 
To see that let us consider the trivial case in which $\mathbf S$ 
is diagonal, i.e. there are no similar features. In this case the input and the latent spaces are equivalent. 
Under such setting the regulariser will have no effect since there are no similarity constraints to 
impose on the model. If on the other hand, all features are identical, i.e.  $S_{ij}=1, \forall i,j$, 
then the latent space will have a dimensionality of one, in such a case the regulariser has the 
strongest effect. 

To explore this dependency we generate two additional synthetic datasets where
we use the same generating mechanism as in A1 but vary the proportion of features we
cluster together to generate latent factors.  Concretely in the synthetic dataset we will call A2 
we randomly select a set A of $d/2$ features over which we will perform clustering to define latent 
factors.  We use the remaining set B of $d/2$ features directly as they are in the latent space.
We cluster the features of the A set to $d/4$ clusters---latent factors, making sure that as in A1 each cluster
has at least one feature in it. As a result the final latent space has a dimensionality of $d/4+d/2=3d/4$. 
To generate the class assignments we proceed as in A1.
To generate the third dataset, A3, we select $d/4$ features to generate A and the remaining for B. We now cluster 
the features in A to $d/8$ clusters, again making sure that there is at least one feature pre cluster. The 
dimensionality of the latent space is now $d/8+3d/4=7d/8$; class assignments are generated as above. 
We used the same values for $n$, $d$, $q$ as in A1.
As we move from A1 to A3 we reduce the number of features that are similar to other features, thus we 
increase the sparsity of $\mathbf S$. For A2 and A3 the percentage of non-zero elements is 
$0.021\%$ and $ 0.011\%$ respectively, compared to $0.04\%$ we had in A1. So A1 is the datasets that has
most constraints while A3 is the one with the least constraints. We apply the different regularisers
in these two datasets using exactly the same protocol as in A1. The results are also given in table~\ref{artificial2}.
As we see the classification error of both ST and AN increases as the dataset sparsity increases and it approaches
that of the standard regularisers.

\note[Removed]{
In order to explore how the sparsity level of $S$ effects the performance of our regualarizer, 
we have repeated the artificial data generation process twice more, but this time we only cluster $\frac{d}{2}$ 
and $\frac{d}{4}$ of the original features to $\frac{d}{4}$ and $\frac{d}{8}$ groups respectively, and
leave the rest of the features as it is and use them as each of them alone form a single cluster. This 
gives us  $\frac{d}{4}+\frac{d}{2}$ and  $\frac{d}{8}+ \frac{d}{2}$ number of latent factor respectively 
for each data set. The rest is followd same as before, By doing so, we generated two more data set which 
have more sparse structured $S$  ($0.021\%$ and $ 0.011\%$ sparsity than the prevously generated one.)}
\note[Amina]{this is actually 
calculated by not counting the diagonal in, but i we say a matris sparsity is this much percent of 
it is total element is not zero, then should we count the diagonals too?  in that case the sparsity would 
be $0.54\%$ and $0.44$ from this two and for previous one is $0.76$}   
\note[Removed]{
We set $d=3000$, $n=5000$, $m=5$ same as before. Applying the same experiment setting as we did for the 
previously generated dataset, the result shows, indeed, The more sparse $S$ are, the weaker the regularizer 
effects are, see the tale \ref{artificial2}. }
\note[Amina]{this is true in this very special artificial data 
set becuse each feature only belong to one class, all weights corrensponding to the features in one class 
pushed towards one same value, but in real data, each features belong to many classes, that means the weigt 
corresponding to such features are pushed twards very different direction at each updata in stochstic gradient 
descent, in batch case, then it is even worse that it is pushed twards different direction at the same time. 
There for in real word problem i believe there is a peak where the regualrzier has best effect with respect 
to sparsity then it drops again, not monotone } 

\begin{table}[h!]
\begin{center}
\scalebox{0.8}{
\begin{tabular}{ r|r|r|r|r|r} 
 \hline
	Dataset & $\{S_{ij} \neq 0\}$&  ST&AN          &$\ell_2$&Dropout \\ \hline
	A1      &     $0.04\%$         & 43.70 &44.00 &52.70    & 53.60 \\ \hline
	A2      &     $0.021\%$      & 50.08 &51.00 &  55.40       & 55.30 \\ \hline
	A3      &     $0.011\%$      & 56.20 & 52.50 &  54.50 & 55.90\\ \hline
\end{tabular}
}
\end{center}
	\caption{Classification error, \%, of the different regularizers, and \% of non zero elements of the similarity matrix $\mathbf S$
	for the three artificial datasets.}
\label{artificial2}
\end{table}

\subsection{Real world datasets}
We evaluated both approaches on the eight classification datasets used in \cite{kusner2015word}. 
The datasets are: BBC sports articles (BBCSPORT) labeled as one of athletics, cricket, footbal, rugby, tennis; 
tweets labeled with sentiments ‘positive’, ‘negative’, or ‘neutral’ 
(TWITTR); recipes labeled by their region of origin (RECIPE); of medical abstracts labeled 
by different cardiovascular disease groups (OHSUMED); sentences from academic papers labeled 
by publisher name (CLASSIC); amazon reviews labeled by product category (AMAZON); news dataset 
labeled by the news topics (REUTER); news articles classified into 20 different categories (20NEWS). 
We removed all the words in the SMART stop word list  \cite{salton1988term}. Documents are represented as bag of words. 
To speed up training, we removed words that appear very few times over all the documents of a dataset. 
Concretely, in 20NEWS we reduce the dictionary size by removing words with a frequency less or equal to three. 
In the OHSUMED and CLASSIC datasets we remove words with frequency one and the in REUTER dataset words with 
frequency equal or less than two. 
\note[Alexandros]{Why you have different cut off frequencies per dataset?}
As feature side-information we use the word2vec representation of the words which have a dimensionality of 300 
\cite{mikolov2013distributed};
other possibilities include knowledge-based side-information, e.g. based on WordNet \cite{miller1995wordnet}. 
\note[Alexandros]{What is the size of your word2vec representations.}
In table \ref{tabel1} we give a description of the final datasets on which we experiment
including the number of classes ($m$) and average number of unique words per document.
\begin{table}[h!]
	\begin{center}
\scalebox{0.8}{
\begin{tabular}{c|c|c|c|c } 
 \hline
 Date set& n &d &Unique words(avg)& $m$ \\ 
  \hline
    BBCsport &590& 9759&80.9&5\\
    \hline
     Twitter & 2486&4076 &6&3\\
     \hline
      Classic & 5675&7628&34.5&4\\
      \hline
       Amazon & 6400& 4502&28.8&4\\
  \hline
 20NEWS &11293& 6859&51.7&20\\
  \hline
 Recipe & 3496&4992&44.7&15\\
  \hline
 Ohsumed &3999&7643&50&10\\
  \hline
 Reuter & 5485& 5939&33&8\\
  \hline  
\end{tabular}
}
\caption{Data set description}
\label{tabel1}
\end{center}
\end{table}

We compute the similarity matrix $\mathbf S$ from the word2vect word representations using the 
heat kernel with bandwidth parameter $\sigma$, i.e. the similarity of $i,j,$ features is given by: 
$S_{ij}=\exp(-\frac{1}{2\sigma^2}(\mathbf {z}_i-\mathbf z_j)^T(\mathbf z_i-\mathbf z_j))$.
We select $\sigma$ so that roughly 20\% of the entries of the similarity matrix are in $[0.8, 1]$ 
interval.
\note[Amina]{20\% of pair distances of the features?? because S is symmetric, i actually set the value of sigma such that  to 
big 20\% of $d(z_1,z_2)$ is in [0.8,1]}
\note[Alexandros]{Did not understand that...}

For those datasets that do not come with a predefined train/test split (BBCSPORT, TWITTER, CLASSIC, AMAZON, RECIPE), 
we use five-fold cross validation and report the average error. We compare the statistical significance of the results
using the MacNemar's test with a significance level of 0.05. For hyperparameter tuning we use 
three-fold inner cross validation.  We select the $\lambda$ hyperparameters of 
of AN, ST, and  $\ell_2$ from  $[0.001,0.01,0.1,1,10]$; we select the $\lambda$ of dropout 
from  $[0.1,0.2,0.3,0.4,0.5]$. We do a series of experiments in which we vary the number of hidden 
layers. Due to the computational complexity of the backprogation for the AN regulariser we only 
give results for the single layer architecture.

In the first set of experiments we use a neural network with one hidden layer and 100 hidden units, we give the 
results in table~\ref{singleLayer}. ST is significantly 
better than the AN in five out of the eight datasets, significantly worse once, and equivalent in one
dataset. ST is significantly better than the $\ell_2$ in six out of the eight daasets, while it is equivalent in one.
Compared to dropout it is four times significantly better and three times significantly worse.

When we increase the number of hidden layers to two with 500 and 100 units on the first and second layer ST
method is significantly better compared to $\ell_2$ three times, significantly worse three times, while 
there is no significant difference in two datasets. A similar picture emerges with respect to Dropout
with ST being significantly better three times, significantly worse twise, while in three cases 
there is no significant difference. We give the detailed results in table~\ref{twoLayers}.

\note[Alexandros]{Definetely additional experimentation is needed. That means either different types of side-information
such as the ones you can get for WordNet and/or new datasets. In addition we should get all these results also with the
Firmenich datasets. Finally if for the results you have so far, do you think we can try to do better tunning since 
you did not have so much time to do that?}

\begin{table}[h!]
\begin{center}
\scalebox{0.8}{
\begin{tabular}{ r|r|r|r|r|r} \hline
Dataset	 & ST                  &    AN    &$\ell_2$& Dropout        & WMD \\ \hline
BBCsport &  3.39-=-            &        {2.17}==   &  2.72= &         {2.17} & 4.6       \\    \hline
Twitter  &         { 26.90}+++ &31.18=-   & 31.18- &28.44           & 29.00\\         \hline
Classic  &            3.54+++  & 4.03+=   &  5.13- & 3.98           &        {2.80}\\ \hline
Amazon   &        { 6.25}++-   & 7.80=-   &  7.57- & 6.44           & 7.40 \\ \hline
20NEWS   &        {19.58}+++   &23.75=-   & 23.21- &21.31           &27.00 \\ \hline
Recipe   &        {38.76}+++   &43.14--   & 41.21- &39.88           &43.00\\  \hline 
Ohsumed  &     34.45+==       &35.75   =-      & 35.26 = &       {34.39}  &44.00\\  \hline
Reuter   &          3.84=+-    &3.38+=    &  6.03- &        {3.2}   & 3.50\\  \hline
\end{tabular}
}
\caption{Classification error, \%, with one hidden layer NNs. AN: analytical approach, ST: 
stochastic approach.  WMD results are from
from \cite{kusner2015word}.  The $+,-$ and $=$  signs give the significance test results of the 
comparison of the performance of a given regulariser to those of the regularisers in the subsequent 
columns. With $+,-,=$ indicating respectively
significantly better, worse, no difference. WMD is not included in the significance comparison
since at the time of the experiments we did not have acces to the code.}
\label{singleLayer}
\end{center}
\end{table}

\note[Removed]{
\begin{table}[h!]
\begin{center}
\scalebox{0.8}{
\begin{tabular}{ r|r|r|r|r} \hline
Dataset	 & ST        &AN      &$\ell_2$& Dropout   \\ \hline 
BBCsport &   3.39-=- & 2.17== &  2.72= & 2.17      \\ \hline 
Twitter  &  26.90+++ &31.18=- & 31.18- &28.44       \\ \hline 
Classic  &   3.54+++ & 4.03+= &  5.13- & 3.98       \\ \hline 
Amazon   &   6.25++- & 7.80=- &  7.57- & 6.44       \\ \hline 
20NEWS   &  19.58+++ &23.75=- & 23.21- &21.31       \\ \hline 
Recipe   &  38.76+++ &43.14-- & 41.21- &39.88       \\ \hline 
Ohsumed  &  34.45+== &35.75=- & 35.26= &34.39       \\ \hline 
Reuter   &  3.84=+-  &3.38+=  &  6.03- & 3.20       \\ \hline 
\end{tabular}
}
\caption{Classification error, \%, with one hidden layer NNs. AN: analytical approach, ST: 
stochastic approach.  
	The $+,-$ and $=$  signs give the significance test results of the 
comparison of the performance of a given regulariser to those of the regularisers in the subsequent 
columns. With $+,-,=$ indicating respectively
significantly better, worse, no difference. 
	}
\label{singleLayer}
\end{center}
\end{table}
}

\begin{table}[h!]
\begin{center}
\scalebox{0.8}{
\begin{tabular}{r|r|r|r|r} 
 \hline
Dataset	  &    ST  &$\ell_2$& Dropout      & WMD \\ \hline
BBCsport  & 2.04=+ &  2.85= & 2.99      &4.6\\ \hline
Twitter   &27.93-- & 26.64= &26.74   &29.00\\ \hline
Classic   & 3.71+= &  4.51- & 3.69          &2.8\\ \hline
Amazon    & 5.96++ &  7.49- & 6.75         &7.40 \\ \hline
20NEWS    &20.72++ & 22.49= &22.48         &27.00 \\ \hline
Recipe    &41.53-- & 39.61= &39.31        &43.00\\ \hline
Ohsumed   &35.03== & 34.95= &35.14          &44.00\\ \hline
Reuter    & 4.39-= &  3.88= & 4.16      &3.50\\ \hline  
\end{tabular}}
\caption{Classification error, \%, with two hidden layers network. Table interpretation as in table~\ref{singleLayer}}
\label{twoLayers}
\end{center}
\end{table}

\note[Removed]{
\begin{table}[h!]
\begin{center}
\scalebox{0.8}{
\begin{tabular}{r|r|r|r|r} 
 \hline
    Dataset	  &    ST           &$\ell_2$& Dropout   & WMD \\ \hline
    BBCsport  &       {1.36} +=&   2.85 - & 1.49&4.6\\ \hline
    Twitter &  29.28 =-     & 28.35- &       {27.35}&29.00\\ \hline
    Classic  &         {3.55}++&4.82-& 3.69&2.8\\ \hline
    Amazon    &         {6.38} ++ & 7.86-         & 6.92 &7.40 \\ \hline
    20NEWS    & 23.35+-   & 25.62- &       {20.68}&27.00 \\ \hline
    Recipe  &42.86 =-& 42.52 -   &       {38.95
   } &43.00\\ \hline
    Ohsumed  & 36.62=-  & 37.09-         &       {34.78} &44.00\\ \hline
    Reuter   &  4.34--& 3.61=&  3.24 &       {3.50}\\ \hline  
\end{tabular}}
\caption{Classification error, \%, with two hidden layers network. Table interpretation as in table~\ref{singleLayer}}
\label{threeLayers}
\end{center}
\end{table}
}

\note[Removed]{

\begin{table}[h!]
\begin{center}
\scalebox{0.8}{
\begin{tabular}{r|r|r|r|r} 
 \hline
Dataset	      &    ST    &$\ell_2$& Dropout   & WMD \\ \hline
    BBCsport  &   1.36+  &  2.72  & 1.49        &  4.60 \\ \hline
    Twitter   &  26.90-  & 26.64  &26.74        & 29.00 \\ \hline
    Classic   &   3.54+  &  4.51  & 3.83        &  2.8 \\ \hline
    Amazon    &   5.96+  &  7.49  & 6.44        &  7.40 \\ \hline
    20NEWS    &  19.58+  & 22.49  &20.68        & 27.00 \\ \hline
    Recipe    &  38.76-  & 39.61  &38.95        & 43.00 \\ \hline
    Ohsumed   &  34.45-  & 34.95  &34.39        & 44.00 \\ \hline
    Reuter    &   3.84-  &  3.61  & 3.20        &  3.50 \\ \hline  
\end{tabular}}
\caption{Best classification error for each algorithm, over all the network architectures.}
\label{bestOfEachAlgo}
\end{center}
\end{table}

}

\note[Alexandros]{Might be worth while examining the significance differences 
in the table~\ref{bestOfEachAlgo}, although reporting the best is not very sound 
methodologically.}

\section{Conclusion and Future Work}
Many real world applications come with additional information describing the properties of the features. Despite that,
quite limited attention has been given to such setting. In this paper we develop a regulariser that exploits exactly
such information for general non-linear models. It relies on the simple intuition that features which have similar 
properties should be treated by the learned model in a similar manner. The regulariser imposes a stability
constraint over the model output. The constraint forces the model to produce similar outputs 
for instances the feature values of which differ only on similar features.
We give two ways to approximate the value of the regulariser. An analytical one which boils
down to the imposition of a Laplacian regulariser on the Jacobian of the learned model with respect to the input features
and a stochastic one which relies on sampling.

We experiment with neural networks with the two approximations of the regulariser and compare 
their performance to well established model regularisers, namely $\ell_2$ and dropout, on artificial and real world 
datasets. In the artificial datasets, for which we know that they match the assumptions of our regulariser we
demonstrate significant performance improvements. In the real world datasets the performance improvements are
less striking. One of the main underlying assumptions of our model is that the feature side-information is 
indeed relevant for the learning problem, when this is indeed the case we will have performance improvements.
If it is not the case then the regulariser will not be selected, as a result of the tuning of the $\lambda$
parameter. 
\note[Amina]{in the experiment , during cross validation to tune the parameter, some of the time 
(a lot actually ) the best result happen at lambda=0,   later we decided to remove  lambda=0 from 
cross validation,  on those data set, it happened most of the time even just standard net do better 
than any of them with a regularizer,  i feel like this sentence will rise a lot of question }
\note[Alexandros]{Hhmmmm... now this is an issue... so that means that for these datasets regularisation
is useless?!?!?!?! problem...}
Along the same lines we want to perform a more detailed study on how the structure of the similarity 
matrix, namely its sparsity and the underlying feature cluster structure, determines the regularisation strength 
of our regulariser. It is clear that a sparse similarity matrix will lead to a rather limited regularisation 
effect since only few features will be affected. This points to the fact that the regulariser should be used
together with more traditional sparsity inducing regularisers, especially in the case of a sparse feature 
feature similarity matrix. Finally since we use the feature information through a similarity function it might 
be the case that the similarity function that we are using is not appropriate and better results can be obtained 
if we also learn the feature similarity. We leave this for future work.

\note[Removed]{
We have presented two approaches that incorporate the feature side-information within learning to constrain 
the model to respect feature similarity. Both of them are inspired by the same intuition: features with similar 
side-information should have similar effect on the learned model output. such idea formulated by the assumption that 
changing the relative contribution of two similar input features, while keeping their total contribution fixed, 
should have only a small effect on the model's output. The first approach relies on the derivatives of the model 
output w.r.t. the input features. These derivatives measure the model's sensitivity to the input features. We 
apply a Laplacian regulariser on them to force the model's sensitivity to the features reflect the 
feature manifold as the latter is defined by the feature side-information. 

We developed a variant of the backpropagation 
algorithm that allows the computation of the gradient of cost functions that include the derivatives of the model 
with respect to the input (Jacobian of the learned model). This modified backpropagation algorithm will help us 
avoid directly calculation of the Hessian during optimization, however, the propagation terms involve calculation of 
tensors which again computationally expensive compare to the standard backpropagation algorithm. Therefore, we propose 
a second variant that makes use of data augmentation to implement the same intuition and has a 
simpler cost function. We generate instances by using the feature similarity matrix as this is computed from the 
feature side-information.  We performed experiments on a set of document classification datasets which show important 
performance gains with respect to standard regularisers as well as WMD which uses feature side-information. One basic 
assumption of the proposed learning models is that the feature side-information is relevant for the given learning 
tasks and they can be used as is. Feature similarity matrix is calculated independent of the model which assume that 
the similarity is correct similarity with respect to the task at hand, however it is true most of the cases as long 
as we make a reasonable similarity according to the relation of features property and task at hand. For future work, 
we are more interested to learn the similarity measure given the side-information together with the prediction function 
which will automatically learn a reasonable measure with respect to the task at hand.
}

\newpage

\bibliographystyle{icml2017}
\bibliography{ms}

\clearpage

\section{Appendix}
\subsection{Modified Backpropogation}
For notion simplicity, we consider stochastic gradient descent. The objective function we want to minimize is as following:
\begin{equation}\label{derivative model2}
E= L(\mathbf y,\vect \phi(\mathbf x))+\lambda_1 \sum_{ij} || \frac{\partial \vect \phi(\mathbf x)}{\partial x_i}  -  \frac{\partial \vect \phi(\mathbf x)}{\partial x_j} ||^2 S_{ij}
\end{equation}
Notice that the objective function includes derivative of the learned function with respect to the input features, if we use neural network to learn the model, the conventional backpropagation algorithm can't be applied directly. Therefore, we developed a modified version of the backpropagation algorithm to find the gradient of the objective.\\
We keep the notation consistent with the notation used in the book of \cite{bishop1995neural}. n is the total layers (including input and out put layer) number of the network, $a^k$ is the pre-activation units in layer $k$, $k_1$ is the number of hidden units in hidden layer $k$, $m$ is the number of output units, and $h(x)$ stands for the non-linear activation function. 
\begin{eqnarray}
\begin{split}
\mathbf z^0=\mathbf x\\
\mathbf a^k=\mathbf w^k \mathbf z^{k-1}+\mathbf b^k\\
\mathbf z^k=h(\mathbf a^k)\\
\vect \phi(\mathbf x)=\mathbf z^n
\end{split}
\end{eqnarray}
To find the gradient of \eqref{derivative model2}, we define $\mathbf{\delta}^k$ as the Jacobian of the learned function with respect to pre-activations at the layer $k$:
\begin{equation}
\vect\delta^k=\begin{bmatrix}
\frac{\partial \vect \phi_1}{\partial a^k_1}&\frac{\partial \vect \phi_2}{\partial a^h_1}&\cdots.&\frac{\partial  \vect\phi_m}{\partial a^k_1}\\
\frac{\partial \vect \phi_1}{\partial a^k_2}&\frac{\partial \vect \phi_2}{\partial a^k_2}&\cdots&\frac{\partial\vect \phi_m}{\partial a^k_2}\\
\vdots&\vdots&\vdots\\
\frac{\partial  \vect\phi_1}{\partial a^k_{k_1}}&\frac{\partial \vect \phi_2}{\partial a^k_{k_1}}&\cdots&\frac{\partial \vect \phi_m}{\partial a^k_{k_1}}
\end{bmatrix}
\end{equation}
 $\vect\delta^k$ for all $k$ can be achieved by the following backpropagation equation.
 \begin{equation}\label{nonlinearRegularizer3}
\vect\delta^k=((\mathbf{W}^{k+1})^T\vect\delta^{k+1})\odot h'(a^k) \:\:\:\forall k=1,2,...,n-1
\end{equation}
Where $\odot $ stands for the element wise multiplication of a column vector to every column of the matrix.
\begin{equation}
\vect\delta^n=\begin{bmatrix}
h'(a^n_1)&0&...&0\\
0&h'(a^n_2)&...&0\\
...&...&...\\
0&0&...&h'(a^n_m)
\end{bmatrix}
\end{equation}
Defining the term $\vect\delta$ in such a away, we can rewrite the regularizer term in equation \eqref{derivative model2} as following:
\begin{equation}\label{reg2}
\sum_{ij} || (\mathbf W^{1}(:,i))-\mathbf W^{1}(:,j))^T\vect\delta^1||^2 S_{ij}
\end{equation}
If the network only has one hidden layer, we can derive  derivative of the regularizer with respect to weights using $\vect\delta $ and \eqref{nonlinearRegularizer3}. When hidden layer's number is more than one, we need to introduce two more term, one to the backward path and one to the forward path:
Define $\mathbf G^k$ as the jacobian of pre-activation unit at layer $k$ with respect to pre-activation at first hidden layer, note layer $k=1$ corresponding to first hidden layer. 
\begin{equation}
G^k_{mg}=\frac{\partial a^k_m }{\partial a^1_g} \:\:\:\forall k=1,2,3,...,n
\end{equation}
We know that:
\begin{equation}
G^1_{mg}=\frac{\partial a^1_m }{\partial a^1_g}= \left\{
  \begin{array}{lr}
  1 \:\:\: \text{if m=g}\\
  0 \:\:\: \text{others}
  \end{array} \right.
\end{equation}
And $\mathbf G^k$ for all $k$ can be achieved during forward path by the following forward propagation equation and $\mathbf G^1$
\begin{equation}
G^k_{mg}=\sum_l W^k_{ml}G^{k-1}_{lg}h'(a^{k-1}_l) \:\:\:\forall k=2,3,...,n
\end{equation}

Define $\mathbf B^k$ which gives the derivative of the $\vect\delta^k$ with respect to the pre-activation units in the first hidden layers:
\begin{equation}
B^k_{ljg}=\frac{\partial \delta^k_{lj}}{\partial a^1_g}\:\:\:\forall k=1,2,...,n
\end{equation}
We know that: 
\begin{equation}
B^n_{ljg}=\frac{\partial \delta^n_{lj}}{\partial a^1_g}=h''(a_l^n)\mathbf 1_{lj}G^n_{lg}
\end{equation}

$\mathbf B^k$ for all $k$ can be obtained by the following propagating equation during backward path using $\mathbf B^n$ as following:

 \begin{eqnarray}
 \begin{gathered}
\scalebox{0.80}[1]{$
B^k_{ljg}=h''(a_l^k)G^k_{lg}\sum_p \delta^{k+1}_{pj}W^{k+1}_{pl}+h'(a_l^k)\sum_pW^{k+1}_{pl}B^{k+1}_{pjg}
 $}
 \end{gathered}
 \end{eqnarray}
 
\[\forall k=1,2,...,n-1\]

Finally, the gradient of the regularizer, i.e. second term of the equation \eqref{derivative model2}, can be calculated as follwoing:\\
For $k=1$, i.e. first hidden layer:
\begin{eqnarray}
 \begin{gathered}
\scalebox{0.60}[1]{$
\frac{\partial R}{\partial W^1_{lm}}=
4\lambda_1 \sum_{s}S_{ms}\sum_j (\mathbf W^1(:,m)-\mathbf W^1(:,s))^T\vect\delta^1(:,j) \delta^1(lj) $}\\
\scalebox{0.70}[1]{$
+2\lambda_1 \sum_{ks} S_{ks}\sum_j (\mathbf W^1(:,k)-W^1(:,s))^T\vect\delta^1(:,j) \sum_g (W^1(g,k)-W^1(g,s))B^1_{ljg}z^0_m)$}
  \end{gathered}
\end{eqnarray}
For $k=2,...,n$:
\begin{eqnarray}
 \begin{gathered}
\scalebox{0.80}[0.9]{$
\frac{\partial R}{\partial W^k_{lm}}=2\lambda_1 \sum_{ks} S_{ks}\sum_j (\mathbf W^1(:,k)-\mathbf W^1(:,s))^T\vect\delta^1(:,j) $}\\
\scalebox{0.70}[0.9]{$
 \sum_g (W^1(g,k)-W^1(g,s))(z_m^{k-1}B^k_{ljg}+\delta_{lj}^k h'(a^{k-1}_m) G_{mg}^{k-1}) 
 $}
  \end{gathered}
\end{eqnarray}

Gradient with respect to bias term, for all $k=1,...,n$:
\begin{eqnarray}
 \begin{gathered}
\frac{\partial R}{\partial b^k_{lm}}=2\lambda_1 \sum_{ks} S_{ks}\sum_j (\mathbf W^1(:,k)-\mathbf W^1(:,s))^T\vect\delta^1(:,j) \\
 \sum_g (W^1(g,k)-W^1(g,s))B^k_{ljg}
\end{gathered}
\end{eqnarray}
The gradient of the first part of the objective which is some loss function we chose, is same as in the standard Backpropagation algorithm, here we just need to rewrite it in terms of the newly defined $\vect\delta$. For example, if we use sigmoid on all layers as activation function and cross entropy loss, we have the following:
\begin{equation}
\begin{aligned}
&E=-\sum_{i=1}^m (y_{i}\log{\phi(x)_i}+(1-y_{i})\log(1-\phi(x)_i))\\
\end{aligned}
\end{equation}
\begin{equation}
\frac{\partial E}{\partial \mathbf W^k}=\vect\delta^k \frac{\vect\phi-\mathbf y}{\vect\phi (1-\vect\phi)}(\mathbf z^{k-1})^T
\end{equation}
\begin{equation}
\frac{\partial E}{\partial \mathbf b^k}=\vect\delta^k \frac{\vect\phi-\mathbf y}{\vect\phi (1-\vect\phi)}
\end{equation}
Now we can find the gradient of the loss with respect to weights in all layers. Compared to the conventional back propagation algorithm, except we have $\vect\delta$ term which is defined differently than the conventional backprop algorithm, we have one more extra term $\mathbf B^k$ to add to the backward path and one more term $\mathbf G^h$ to the forward path. \\

\note[Removed]{
\subsection{Computational complexity}
For one hidden layer network, when stochastic gradient is used:
computational complexity for calculating gradient of the regularizer with respect to weight matrix is $\mathcal{O}(ohd^2)$, $o$ is the output dimention, $h$ is the hidden unit number and $d$ is the feature dimension. After making $S$ matrix sparse, the computational complexity decrease to  $\mathcal{O}(ohd)$. For network with more hidden layers, the complexity of calculating the gradient with respect to weight matrix $W^k$ is $d(\mathcal{O}(h_1h_{k+1}h_kO)+\mathcal{O}(h_kh_{k-1}h_k)+ \mathcal{O}(dh_1h_kO))$ where $h_k$ refers to the number of hidden neurons in the hidden layer $k$. After sparsifying the similarity matrix $S$ we reduce the computational complexity by factor $d$ and it reduced to $\mathcal{O}(h_1h_{k+1}h_kO)+\mathcal{O}(h_kh_{k-1}h_k)+\mathcal{O}(dh_1h_kO)$
}

\end{document}